\documentclass[10pt,twocolumn,letterpaper]{article}

\usepackage[pagenumbers]{cvpr}

\usepackage{amsmath}
\usepackage{amssymb}
\usepackage{booktabs}
\usepackage{multirow}
\usepackage{graphicx}
\usepackage{xcolor}

\definecolor{cvprblue}{rgb}{0.21,0.49,0.74}
\usepackage[pagebackref,breaklinks,colorlinks,allcolors=cvprblue]{hyperref}

\def\paperID{*****}
\def\confName{CVPR}
\def\confYear{2026}

\title{Solution for 10th Competition on Ambivalence/Hesitancy (AH) Video Recognition
Challenge using Divergence-Based Multimodal Fusion}

\author{
Aislan Gabriel O. Souza, Agostinho Freire, Leandro Honorato Silva, \\Igor Lucas B. da Silva, João Vinícius R. de Andrade, Gabriel C. de Albuquerque,\\Lucas Matheus da S. Oliveira, Mário Stela Guerra, Luciana Machado\\
Universidade de Pernambuco (UPE)\\
Escola Politécnica de Pernambuco (POLI)\\
Recife, Brazil\\
{\tt\small aislan.gabriel@upe.br, aafj@ecomp.poli.br, leandro.ssilva@upe.br, ilbs@poli.br,}\\
{\tt\small jvlc@ecomp.poli.br, gca@ecomp.poli.br, lmso@ecomp.poli.br, msg4@poli.br, lml@ecomp.com.br}
}

\begin{document}
\maketitle

\begin{abstract}
We address the Ambivalence/Hesitancy (A/H) Video Recognition Challenge at the 10th ABAW Competition (CVPR 2026). We propose a divergence-based multimodal fusion that explicitly measures cross-modal conflict between visual, audio, and textual channels. Visual features are encoded as Action Units (AUs) extracted via Py-Feat, audio via Wav2Vec 2.0, and text via BERT. Each modality is processed by a BiLSTM with attention pooling and projected into a shared embedding space. The fusion module computes pairwise absolute differences between modality embeddings, directly capturing the incongruence that characterizes A/H. On the BAH dataset, our approach achieves a Macro F1 of 0.6808 on the validation test set, outperforming the challenge baseline of 0.2827. Statistical analysis across 1{,}132 videos confirms that temporal variability of AUs is the dominant visual discriminator of A/H.
\end{abstract}

\section{Introduction}

Ambivalence and hesitancy (A/H) are complex affective states characterized by conflicting emotions or intentions toward behavioral change~\cite{gonzalez-25-bah, miller2012motivational}. Unlike discrete emotions~\cite{li2020deep}, A/H manifests as \emph{incongruence} across communication channels. We hypothesize that this cross-modal conflict can be explicitly modeled in multimodal fusion.

Multimodal affective computing has made significant progress~\cite{poria2017review, baltrusaitis2019multimodal, gandhi2023multimodal}, yet standard fusion strategies---concatenation~\cite{tzirakis2017end2end} or late blending~\cite{savchenko2025hsemotion}---treat modalities as complementary rather than potentially conflicting. For A/H detection, where the signal lies precisely in the \emph{disagreement} between what a person says, how they sound, and what their face shows, we argue that measuring cross-modal divergence is more informative.

We propose a divergence-based fusion strategy that computes pairwise absolute differences between projected modality representations. For visual features, we extract interpretable Action Units (AUs) via Py-Feat~\cite{cheong2023pyfeat} and encode temporal dynamics through sliding window statistics. Audio features are extracted via Wav2Vec 2.0~\cite{baevski2020wav2vec} and text via BERT~\cite{devlin2019bert}.

Our contributions include: (1) a divergence-based fusion that explicitly models cross-modal conflict; (2) temporal windowed AU representations capturing facial instability; (3) statistical analysis showing that AU variability, not mean intensity, discriminates A/H; and (4) a complete ablation across three modalities and three fusion variants on the BAH dataset~\cite{gonzalez-25-bah}.

\section{Related Work}

The ABAW competition series~\cite{kollias2025abaw8, kollias2024abaw6, zafeiriou2017affwild, kollias2019affwild2} has driven progress in affective behavior analysis using in-the-wild datasets such as Aff-Wild2~\cite{kollias2019affwild2} and AffectNet~\cite{mollahosseini2017affectnet}. Recent editions introduced the A/H recognition challenge based on the BAH dataset~\cite{gonzalez-25-bah}.

At ABAW-8, Savchenko~\cite{savchenko2025hsemotion} achieved top performance using EmotiEffLib~\cite{savchenko2024emotiefflib} facial descriptors with Wav2Vec 2.0 audio and RoBERTa~\cite{liu2019roberta} text embeddings trained on GoEmotions~\cite{demszky2020goemotions}, employing late fusion via blending. Hallmen~\etal~\cite{hallmen2025semantic} used ViT-Huge for vision, Wav2Vec 2.0 pre-trained on emotional speech~\cite{wagner2023wav2vec} for audio, and GTE for text, finding that text provides the strongest signal. Both employed standard fusion without explicitly modeling cross-modal conflict. Richet~\etal~\cite{richet2024textualized} proposed the BAH baseline using co-attention over TCN-processed features. Related work in EMI estimation~\cite{hallmen2024unimodal, yu2024efficient} has also explored multimodal fusion for affective tasks.

Multimodal fusion strategies range from early concatenation to late blending and attention-based approaches~\cite{liu2023multimodal, nagrani2021attention, zhang2022continuous}. Our divergence-based approach is architecturally simpler and theoretically motivated by the definition of A/H.

Action Units provide interpretable facial representations and have been modeled with deep learning~\cite{li2018eac, shao2021jaa}. While deep embedding approaches~\cite{savchenko2024emotiefflib} dominate recent competitions, AUs enable analysis of which specific facial behaviors associate with A/H.

\section{Method}

\subsection{Feature Extraction}

\paragraph{Visual.} We extract 20 AUs per frame using Py-Feat~\cite{cheong2023pyfeat} from pre-cropped face images, sampling 1 in 3 frames ($\sim$10\,fps). For temporal modeling, we compute four statistics (mean, std, slope, range) per AU within sliding windows of $W{=}16$ frames (step $S{=}8$), yielding 80-dimensional window descriptors.

\paragraph{Audio.} Audio is extracted at 16\,kHz mono and processed with Wav2Vec 2.0 (\texttt{wav2vec2-base-960h})~\cite{baevski2020wav2vec}, producing 768-dimensional embeddings at $\sim$50\,Hz. Alternative audio models such as HuBERT~\cite{hsu2021hubert} and Whisper~\cite{radford2023whisper} exist but we use the standard Wav2Vec 2.0 for simplicity.

\paragraph{Text.} Transcripts are encoded with BERT-base~\cite{devlin2019bert}, using the [CLS] token (768 dims). The last two BERT layers are fine-tuned with reduced learning rate.

\subsection{Temporal Modeling and Fusion}

Each temporal modality is processed by a 2-layer BiLSTM(hidden dim 64) with attention pooling, then projected to $D{=}128$ dimensions. We compare three fusion strategies:

\textbf{Fusion A (Implicit):} $\mathbf{f}_A = [\mathbf{h}'_v;\, \mathbf{h}'_a;\, \mathbf{h}'_t]$

\textbf{Fusion B (Divergence):} $\mathbf{f}_B = [|\mathbf{h}'_v - \mathbf{h}'_a|;\, |\mathbf{h}'_v - \mathbf{h}'_t|;\, |\mathbf{h}'_a - \mathbf{h}'_t|]$

\textbf{Fusion C (Combined):} $\mathbf{f}_C = [\mathbf{f}_A;\, \mathbf{f}_B]$

The fused vector is classified by a 3-layer MLP with dropout ($p{=}0.3$). Training uses BCEWithLogitsLoss with class weighting, AdamW with differentiated learning rates ($5{\times}10^{-5}$ for BERT, $5{\times}10^{-4}$ for other parameters), cosine annealing over 30 epochs, gradient clipping at 1.0, and early stopping with patience 8.

For unimodal baselines, we also evaluate XGBoost on temporally vectorized features with grid search over hyperparameters.

\section{Experiments}

\subsection{Dataset}

We evaluate on the BAH dataset for the 10th ABAW Competition. After excluding participants without consent, our dataset contains 1{,}132 videos (598 train, 107 val, 427 test) from 239 participants. The metric is Macro F1 at the video level.

\subsection{Statistical Analysis}

Mann-Whitney U tests across 1{,}132 videos reveal that temporal variability (std) of AUs dominates as a discriminator of A/H (Table~\ref{tab:stats}). The top features by effect size are AU06 std ($|r|{=}0.186$), AU09 std ($|r|{=}0.186$), and AU12 std ($|r|{=}0.172$), all significant after Bonferroni correction. This confirms that A/H manifests as facial instability---increased fluctuation of AUs over time---not as a distinct expression.

\begin{table}[t]
\centering
\small
\caption{Top discriminative AU features (Mann-Whitney U, $N{=}1{,}132$). All significant after Bonferroni correction.}
\label{tab:stats}
\begin{tabular}{@{}llcc@{}}
\toprule
Feature & Metric & A/H vs No A/H & $|r|$ \\
\midrule
AU06 (cheek raiser) & std & 0.076 vs 0.059 & 0.186 \\
AU09 (nose wrinkler) & std & 0.095 vs 0.084 & 0.186 \\
AU12 (smile) & std & 0.089 vs 0.068 & 0.172 \\
AU26 (jaw drop) & zcr & 0.421 vs 0.384 & 0.168 \\
AU02 (outer brow) & std & 0.110 vs 0.102 & 0.149 \\
\bottomrule
\end{tabular}
\end{table}

\subsection{Results}

Table~\ref{tab:results} presents unimodal and multimodal results. Audio is the strongest unimodal modality (0.6141), consistent with prior findings~\cite{hallmen2025semantic, hallmen2024unimodal}. Fusion B (divergence) achieves the best test F1 of 0.6808, supporting our hypothesis that cross-modal conflict is a primary indicator of A/H.

\begin{table}[t]
\centering
\small
\caption{Results on the BAH dataset (Macro F1).}
\label{tab:results}
\begin{tabular}{@{}lcc@{}}
\toprule
Model & Val F1 & Test F1 \\
\midrule
\textit{Unimodal:} & & \\
\quad Visual AUs (XGBoost) & 0.6194 & 0.5642 \\
\quad Audio Wav2Vec (LSTM) & 0.5218 & 0.6141 \\
\quad Text BERT & 0.5758 & 0.5904 \\
\midrule
\textit{Multimodal (raw AUs):} & & \\
\quad Fusion A (implicit) & 0.6788 & 0.6604 \\
\quad Fusion B (divergence) & 0.6524 & \textbf{0.6808} \\
\quad Fusion C (combined) & 0.6700 & 0.6766 \\
\midrule
\textit{Multimodal (windowed AUs):} & & \\
\quad Fusion B (divergence) & 0.6912 & 0.6602 \\
\midrule
Challenge baseline~\cite{gonzalez-25-bah} & --- & 0.2827 \\
\bottomrule
\end{tabular}
\end{table}

\section{Discussion and Conclusion}

The success of Fusion B aligns with the theoretical definition of A/H as coexisting conflicting signals. By providing the classifier only with cross-modal divergence information, Fusion B captures the essence of ambivalence more directly than concatenation-based approaches. All visual-only models plateau at ${\sim}0.56$ F1, consistent with the small effect sizes ($|r|{<}0.2$) found in our statistical analysis. The multimodal fusion lifts performance to 0.68 by combining this weak visual signal with stronger audio and text modalities.

A recurring finding is that richer feature representations improve validation but not test performance, reflecting the limited training set (598 videos). Future work should explore temporal alignment between visual and audio within the divergence framework, data augmentation, and the use of additional visual representations such as MediaPipe blendshapes~\cite{lugaresi2019mediapipe} with larger datasets.

{\small
\bibliographystyle{ieeenat_fullname}
\bibliography{main}
}

\end{document}